\documentclass{article}

\usepackage{microtype}
\usepackage{graphicx}
\usepackage{subcaption}
\usepackage{booktabs} 

\usepackage{hyperref}

\usepackage[accepted]{icml2026}

\usepackage{amsmath}
\usepackage{amssymb}
\usepackage{mathtools}
\usepackage{amsthm}
\usepackage{booktabs} 
\usepackage{multirow} 
\usepackage{graphicx} 
\usepackage{xcolor}   

\usepackage[capitalize,noabbrev]{cleveref}

\theoremstyle{plain}
\newtheorem{theorem}{Theorem}[section]
\newtheorem{proposition}[theorem]{Proposition}

\theoremstyle{definition}
\newtheorem{definition}[theorem]{Definition}
\newtheorem{assumption}[theorem]{Assumption}
\theoremstyle{remark}

\usepackage[textsize=tiny]{todonotes}

\icmltitlerunning{FlexCausal: Flexible Causal Disentanglement via Structural Flow Priors and Manifold-Aware
Interventions}

\begin{document}

\twocolumn[
  \icmltitle{FlexCausal: Flexible Causal Disentanglement via Structural Flow Priors and Manifold-Aware
Interventions}
    
  \begin{icmlauthorlist}

        \icmlauthor{Yutao Jin}{seu}
        \icmlauthor{Yuang Tao}{seu}
        \icmlauthor{Junyong Zhai}{seu}
    \end{icmlauthorlist}

    \icmlaffiliation{seu}{Southeast University, Jiangsu, China}

    \icmlcorrespondingauthor{Yutao Jin}{yutao0212@seu.edu.cn}
    \icmlcorrespondingauthor{Yuang Tao}{yuangtao@seu.edu.cn}
    \icmlcorrespondingauthor{Junyong Zhai}{jyzhai@seu.edu.cn}

]
\printAffiliationsAndNotice{}
\begin{abstract}
Causal Disentangled Representation Learning(CDRL) aims to learn and disentangle low dimensional representations and their underlying causal structure from observations. However, existing disentanglement methods rely on a standard mean-field approximation with a diagonal posterior covariance, which decorrelates all latent dimensions. Additionally, these methods often assume isotropic Gaussian priors for exogenous noise, failing to capture the complex, non-Gaussian statistical properties prevalent in real-world causal factors. Therefore, we propose FlexCausal, a novel CDRL framework based on a block-diagonal covariance VAE. FlexCausal utilizes a Factorized Flow-based Prior to realistically model the complex densities of exogenous noise, effectively decoupling the learning of causal mechanisms from distributional statistics. By integrating supervised alignment objectives with counterfactual consistency constraints, our framework ensures a precise structural correspondence between the learned latent subspaces and the ground-truth causal relations. Finally, we introduce a manifold-aware relative intervention strategy to ensure high-fidelity generation. Experimental results on both synthetic and real-world datasets demonstrate that FlexCausal significantly outperforms other methods.
\end{abstract}

\section{Introduction}
\begin{figure}
    \centering
    \includegraphics[width=\columnwidth]{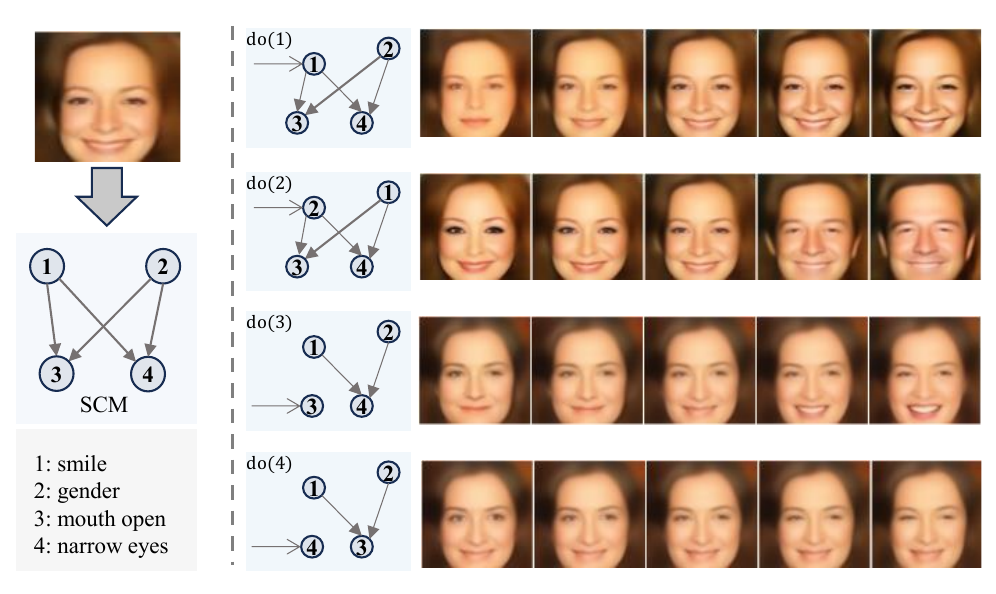}
    \caption{CRDL aims to disentangle latent causal factors from observation and enable controllable causal generation. Our FlexCausal effectively disentangles causal representations. Interventions on parent nodes will influence their child nodes, while interventions on child nodes leave the parent nodes invariant.}
    \label{fig:intro}
\end{figure}

Causality~\cite{pearl2009causality} has significantly advanced modern artificial intelligence, enabling agents not only to observe patterns, but also to comprehend the underlying mechanisms governing the world. Despite unprecedented success in fitting statistical correlations, contemporary deep learning often encounters challenges at higher levels of the "causal ladder", specifically in intervention analysis and counterfactual reasoning.
Unlike traditional Disentangled Representation Learning (DRL) methods, which focus on the statistical independence of latent factors, Causal Disentangled Representation Learning (CDRL)~\cite{scholkopf2021toward,causalvae} shifts the focus toward uncovering the structural causal mechanisms governing high-dimensional semantic concepts.
As demonstrated in Figure~\ref{fig:intro}, by disentangling and controlling these causal representations, deep learning models can facilitate various real-world tasks, ranging from controllable factual synthesis to counterfactual generation. To this end, recent advances have integrated causal semantics into various generative frameworks to enable counterfactual image editing, including VAEs~\cite{causalvae,scmvae}, GANs~\cite{causalgan}, and Diffusion models~\cite{causaldiffae,cidiffuser}.

Despite these advances, existing CDRL frameworks often fail when applied to complex real-world systems, primarily because they fail to mitigate two fundamental limitations. First, they typically rely on the restrictive Gaussian assumption, which fails to capture the complex, non-Gaussian distributions found in real-world data, thereby limiting identifiability. Second, there is a mismatch between the modeled latent structure and the true causal geometry. Standard approaches often ignore the intra-concept block-wise correlations and rely on naive hard interventions that neglect the data manifold. This structural misalignment inevitably results in unrealistic counterfactuals that violate the consistency needed for reliable generation.

Therefore, we propose FlexCausal, a novel CDRL framework aimed at combining structural causal modeling with flexible density estimation. FlexCausal achieves the disentanglement of high-dimensional causal representations by learning independent Normalizing Flows for each concept's exogenous priors, while enforcing strict causal consistency among latent variables during training. It is crucial that FlexCausal facilitates manifold-aware interventions, allowing for counterfactual generation that strictly adheres to the topological structure of each latent concept. By relaxing rigid distributional assumptions and respecting the geometry of data, FlexCausal effectively fosters causal disentanglement and mitigates distributional mismatch in complex real-world scenarios.

Specifically, we introduce a Flow-based Exogenous Prior framework that employs invertible normalizing flows to explicitly model the complex distributions of exogenous variables, effectively relaxing the rigid constraints inherent in traditional causal VAEs like CausalVAE. Furthermore, we propose a Manifold-Aware Directional Intervention mechanism coupled with a Counterfactual Consistency Loss. Different from traditional hard interventions that rely on arbitrary scalar assignments, this mechanism learns semantic directions within a designed Block-Diagonal latent space and performs soft interventions along the natural data manifold. 

Overall, our main contributions are summarized as follows:
\begin{itemize}
\item[$\bullet$]We propose a novel CDRL framework named FlexCausal that synergizes SCMs with deep generative learning via an additive Flow-based prior. By enforcing a strict Block-Diagonal constraint on the posterior distributions and explicitly learning complex exogenous priors, our approach successfully disentangles high-dimensional causal concepts.
\item[$\bullet$]We introduce an additive Flow-based Exogenous Prior that imposes a strict Block-Diagonal constraint on the latent covariance, while simultaneously enforcing a counterfactual consistency constraint during training. This design fundamentally overcomes the theoretical limitations of conventional methods by relaxing rigid Gaussian assumptions and intra-dimensional independence. 
\item[$\bullet$]We propose a novel intervention strategy designed to perceive the direction of the original data manifold. Unlike hard interventions that often yield off-manifold artifacts, our method performs relative interventions along learned semantic directions. This ensures that counterfactual generation maintains high fidelity while strictly adhering to the underlying geometric structure of the data.
\item[$\bullet$]We construct Filter, a new synthetic benchmark where latent concepts are sampled from different complex distributions to simulate real-world environmental complexity. Experimental results demonstrate that our FlexCausal outperforms other methods on both synthetic datasets and standard real-world datasets.
\end{itemize}

\section{Related Works}
\subsection{Disentangled Representation Learning and Causal Representation Learning}
With the development of deep learning, Disentangled Representation Learning (DRL) has emerged as a method capable of extracting independent and interpretable low-dimensional representations from high-dimensional~\cite{whyvae}. The core goal of DRL is to map observations into a latent space where distinct semantic factors are encoded into independent dimensions~\cite{bengio2013representation}. 
Previous works such as $\beta$-VAE~\cite{betavae} and FactorVAE~\cite{duan2022factorvae} achieve this by enforcing statistical independence of latent variables, typically employing a KL divergence penalty mechanism between the posterior distribution and a standard Gaussian prior. However, the strict independence assumption in DRL faces significant theoretical and practical challenges. Theoretically,~\cite{locatello2019challenging} proved that unsupervised disentanglement is impossible without inductive biases. Practically, statistical independence among semantic factors rarely exists in real-world scenarios. To address these limitations, Causal Representation Learning (CRL)~\cite{scholkopf2021toward} focuses on structural causal disentanglement. By integrating SCMs into the representation learning process, CRL aims to uncover causal relationships among latent variables, thereby enabling the recovery of causally related concepts and counterfactual reasoning.
Nevertheless, most existing CRL frameworks~\cite{linearcausal,iCITRIS} are still built upon the restrictive assumption of standard Gaussian priors for exogenous variables. Recent frontier works~\cite{mutirep,mixturecausal} have begun to explore CRL approaches based on mixture priors to better learn the complex, multi-modal causal representation in real-world scenarios.

\subsection{Causal Disentangled Representation Learning}
Different from DRL’s strict assumption of statistical independence, Causal Disentangled Representation Learning (CDRL) instantiates the theoretical principles of CRL by explicitly integrating SCMs into deep generative frameworks. This integration facilitates both robust causal representation learning and controllable causal generation.
CausalVAE~\cite{causalvae}, introduces a specialized Causal Layer that transforms independent exogenous noise into causally endogenous variables via linear structural equations, effectively propagating causal effects during inference. DEAR~\cite{dear} uses a GAN-based architecture combined with SCM priors to enable supervised learning of causal structures. Similarly, SCM-VAE~\cite{scmvae} utilizes additive noise models alongside a fixed SCM to enhance both identifiability and interpretability. ICM-VAE~\cite{icmvae} employs flow-based diffeomorphic mappings to model nonlinear causal mechanisms under the Independent Causal Mechanisms (ICM) principle, achieving theoretically identifiable causal disentanglement. To Address the issue of confounders, CFI-VAE~\cite{cfivae} learns unbiased causal representations by eliminating data bias and C-Disentanglement~\cite{cdisen} relaxes the traditional assumption of confounder-free data by explicitly incorporating the inductive bias of confounders, thereby improving the performance of disentanglement. Building upon the causal insights of CFI-VAE, CIDiffuser~\cite{cidiffuser} incorporates Diffusion models to generate high quality counterfactual images. 

\subsection{Deep Generative Model and Counterfactual Editing}
With the learned causal representation, Deep Generative Models~\cite{gan, ddpm, scorediffusion} can be employed to generate both factual observations and counterfactual alternatives.~\cite{mira} leverage Hierarchical VAEs to perform a strict Abduction-Action-Prediction procedure, achieving high-fidelity counterfactual generation with hard intervention.~\cite{counterfactualedit,pan2025counterfactual} derive theoretical probability bounds within the Augmented SCM (ASCM) framework to establish Counterfactual Consistency, utilizing it primarily as a theoretical metric to verify if distributions fall within identifiable bounds. 
Distinguishing our work from these methods, FlexCausal reformulates counterfactual consistency from a theoretical constraint into a trainable loss function. By explicitly imposing structural penalties on latent vectors during training, we directly embed causal dynamics into the representation learning process. 

Furthermore, addressing the limitations of hard intervention in previous studies~\cite{semangan,causalflow}, we propose a Manifold-Aware Directional Intervention strategy. This ensures that the generated counterfactuals traverse along the high-density geodesic of the learned manifold rather than being projected into invalid regions, thereby preserving the realism of the original data.

\section{Primaries}
We formulate causal representation learning within a VAE framework that maps the high-dimensional data points $x \in \mathcal{X} \subseteq \mathbb{R}^n$ and associated labels $u \in \mathcal{U} \subseteq \mathbb{R}^m$ to the latent variable $z \in \mathcal{Z} \subseteq \mathbb{R}^d$ ($d \ll n$).
The model is defined by a conditional probabilistic encoder $q_\phi(z|x)$ that infers latent causal factors given both the image and the label, and a generator $p_\theta(x|z)$ that reconstructs the observations.

\begin{assumption}[Causal Factorize]
The latent vector $z$ partitions into disjoint causal vector blocks, denoted as $z = [z_1^\top, \dots, z_K^\top]^\top$. Each sub-vector $z_k \in \mathbb{R}^{d_k}$ ($\sum_{k=1}^K d_k = d$) functions as a coherent semantic unit, representing a high-dimensional node in the underlying causal graph.
\label{causal factorize}
\end{assumption}

\begin{definition}
    A Structural Causal Model (SCM) is a tuple $(\mathbf{Z}, \mathbf{N}, \mathbf{F}, P_{\mathbf{N}})$, where:\\
    (1) $\mathbf{Z} = \{z_1, \dots, z_K\}$ is a set of endogenous latent variables.\\
    (2) $\mathbf{N} = \{n_1, \dots, n_K\}$ is a set of exogenous noise variables.\\
    (3) $\mathbf{F} = \{f_1, \dots, f_K\}$ is a set of deterministic structural functions, where $z_k := f_k(\text{PA}(z_k), n_k)$ and $\text{PA}(z_k) \subseteq \mathbf{Z} \setminus \{z_k\}$ denotes the causal parents of $z_k$.\\
    (4) $P_{\mathbf{N}}$ is a product distribution over the exogenous noise.\\
    \label{scm}
\end{definition}
The causal structure is represented by a Directed Acyclic Graph (DAG) $\mathcal{G}$, where a directed edge $z_j \to z_k$ exists if $z_j \in \text{PA}(z_k)$.

\begin{assumption}[Latent Causal Sufficiency]
    \label{ass:dependent}
    A system is causally sufficient, meaning there are no unobserved latent confounders influencing multiple variables in $\mathcal{Z}$. Consequently, the exogenous noise variables are mutually independent:
    \begin{equation}
       p(n) = \prod_{k=1}^K p(n_k).
    \end{equation}
    This implies that the root nodes of $\mathcal{G}$ are mutually independent, and for any node $z_k$, its dependence on the rest of the graph is fully mediated by its causal parents $\text{PA}(z_k)$.
\end{assumption}

\begin{definition}
    An Additive Noise Model (ANM)~\cite{anm} is defined as a structural causal system where each latent variable $z_k$ is generated as a deterministic function of its causal parents $\text{PA}(z_k)$ superimposed with an independent additive noise term $n_k$:
    \begin{equation}
    \label{eq:anm}
    z_k := f_k(\text{PA}(z_k)) + n_k, \quad k=1, \dots, K
    \end{equation}
    where $f_k(\cdot)$ denotes the nonlinear causal mechanism governed by the causal graph $\mathcal{G}$. In our implementation, the parent sets $\text{PA}(z_k)$ and the dependency structure of $f_k$ are explicitly specified by a fixed prior adjacency matrix $A$.
\label{anm}
\end{definition}
\begin{assumption}[Non-Gaussianity]
    To guarantee causal identifiability beyond the Markov equivalence class~\cite{anm}, we assume the distributions of the exogenous noise $p(n_k)$ are non-Gaussian (e.g., Laplace, Bimodal). This assumption justifies our use of Flow-based priors to capture complex distributional geometries.
\end{assumption}

\begin{proposition}
    Let $\mathcal{G}$ be a Directed Acyclic Graph (DAG) representing the causal structure. Under Definition~\ref{anm}, the mapping $\mathcal{T}: \mathcal{Z} \to \mathcal{N}$ defined by $n_k = z_k - f_k(\text{PA}(z_k))$ is a volume-preserving bijective transformation. Specifically, the absolute determinant of its Jacobian matrix is unity:
    \begin{equation}
        \left| \det \frac{\partial n_k}{\partial z_k} \right| = 1.
    \end{equation}
    This property implies that the log-likelihood in the causal latent space equates to the log-likelihood in the exogenous noise space, i.e., $\log p(z_k) = \log p(n_k)$.
    \label{vpt}
\end{proposition}

\begin{figure*}[t]
    \centering
    \includegraphics[width=\linewidth]{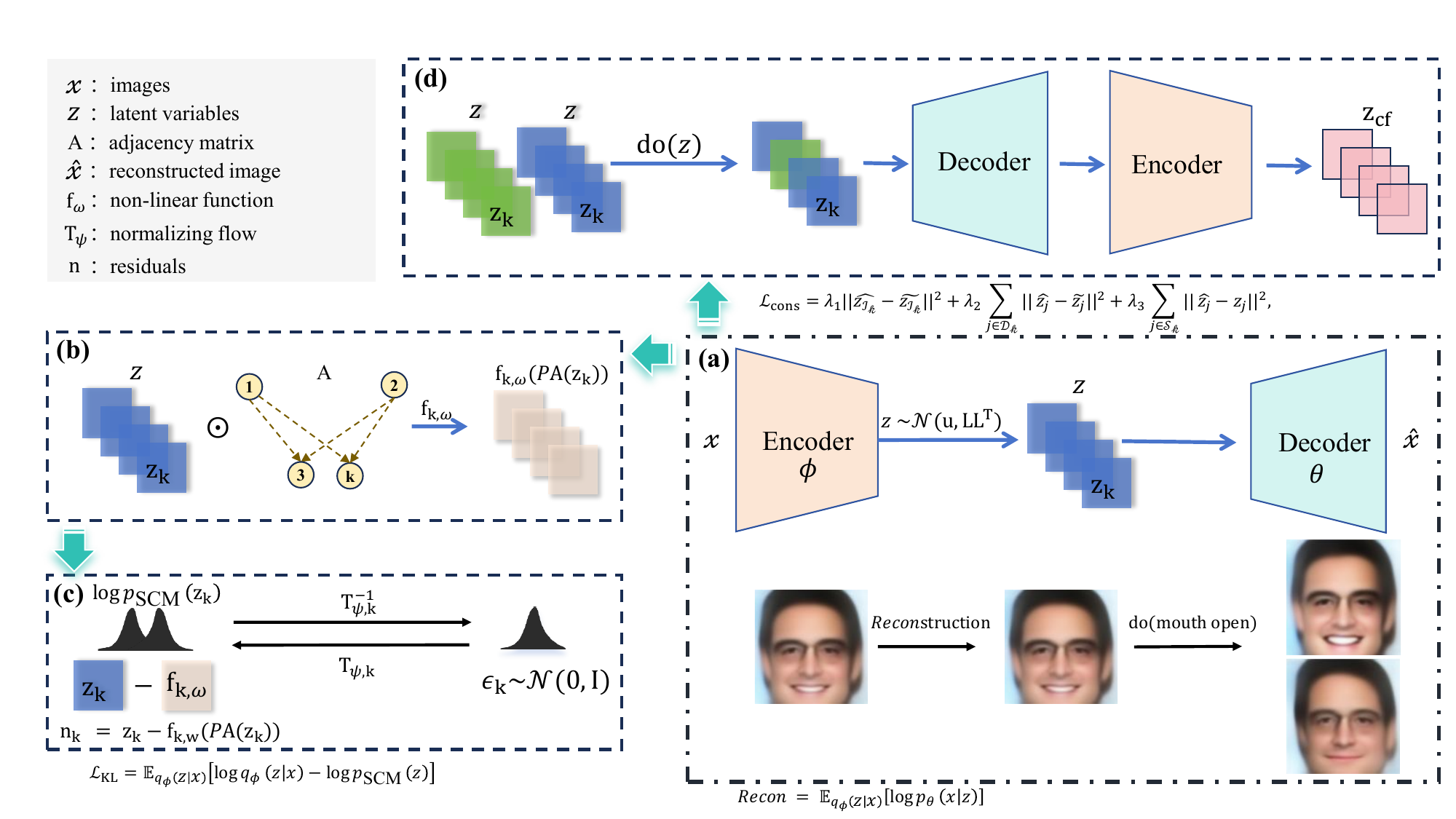}
    \caption{An overview of FlexCausal. (a) Overall Framework: The model encodes input images $x$ into latent variables $z$, which are then reconstructed via the decoder. (b) Structural Causal Mechanism: The latent space is structured as a causal graph defined by the adjacency matrix $A$. For each node $z_k$, its value is determined by its parents $\text{PA}(z_k)$ through a nonlinear structural function $f_{k}$. (c) Flow Prior: To model complex priors, we utilize Normalizing Flows ($T_{\psi, k}$) to transform the simple base distribution into the flexible posterior distribution of the residuals. (d) Counterfactual Consistency Constraint: This explicitly penalizes violations of structural equations under intervention, thereby ensuring the reliability of generated counterfactuals.}
    \label{fig:framework}
\end{figure*}

\section{Methodology}
Figure~\ref{fig:framework} demonstrates the overall architecture of the proposed FlexCausal. Built upon a Block-Diagonal Variational Autoencoder (VAE) backbone, our model encodes observations into a disentangled latent space and captures nonlinear causal relationships among concepts using ANMs. We first derive the exogenous noise residuals $n_k$ from the latent variables $z_k$ using Equation~\ref{eq:anm} and then model their densities with invertible Normalizing Flows. Furthermore, we introduce an intervention-based Counterfactual Consistency Constraint. This mechanism penalizes violations of structural equations under causal intervention, thereby facilitating robust causal disentanglement and ensuring the reliability of counterfactual image generation.
\subsection{Structure-Aware Representation Learning via Block-Diagonal VAE}
Consistent with the ANM framework outlined in Definition~\ref{anm}, the observational data $x$ is generated by a causal latent vector z. The underlying causal topology is pre-defined and encoded by a fixed binary adjacency matrix $A \in \{0, 1\}^{K \times K}$, and $A_{jk} = 1$ indicates the existence of a directed causal edge $z_j \to z_k$. Under this topological constraint, each variable $z_k$ is structurally determined by its causal parents:
\begin{equation}
z_k = f_k\left( { z_j \mid A_{jk} = 1 } \right) + n_k,
\end{equation}
where $z_k \in \mathbb{R}^{d_k}$ represents a distinct vector-valued semantic concept, $f_k(\cdot)$ is a learnable nonlinear structural function, and $n_k$ denotes the independent exogenous noise.

Standard VAEs, which are utilized in CausalVAE and SCM-VAE, typically employ a mean-field approximation. This formulation assumes a diagonal posterior covariance, effectively forcing all latent dimensions to be statistically independent. While this strategy facilitates the disentanglement of distinct concepts, it is detrimental to vector-valued concepts. Crucially, the dimensions within a single concept $z_k$ are inherently coupled. Severing these intra-concept correlations compromises the semantic integrity and geometric structure of the causal variables.

To prevent the destruction of intra-concept correlations, we utilize a Block-Diagonal Variational Encoder $E_\phi$. Our method not only disentangles distinct semantic concepts but also maintains the necessary dependencies within each concept block. We formulate the variational posterior as follow:
\begin{equation}
q_\phi(z|x) = \mathcal{N}(\mu_\phi(x), \Sigma_\phi(x)).
\end{equation}

To achieve the causal factorization postulated in Assumption~\ref{causal factorize}, the covariance matrix $\Sigma_\phi(x)$ is constrained to be block-diagonal. 
Specifically, we require zero correlations between distinct causal concepts $z_k, z_j$ ($k \neq j$) while maintaining dense intra-concept dependencies. We construct $\Sigma_\phi(x)$ as the direct sum of concept-specific covariance blocks:
\begin{equation}
\Sigma_\phi(x) = \text{blkdiag}\left(\Sigma_1(x), \dots, \Sigma_k(x)\right).
\end{equation}
To ensure positive definiteness and numerical stability, we parameterize each block $\Sigma_k$ via its local Cholesky factor $L_k$. The global Cholesky matrix $L$ is composed of these sub-blocks:
\begin{equation}
L = \text{blkdiag}(L_1, \dots, L_K),
\end{equation}
where $L_k = \text{tril}(V_k)$, $V_k \in \mathbb{R}^{d_k \times d_k}$ denotes the raw output sub-matrix from the encoder corresponding to the $k$-th concept, and $\text{tril}(\cdot)$ enforces the lower-triangular constraint. The diagonal elements of each $L_k$ are rectified via softplus. Consequently, the full covariance is given by $\Sigma_k = L_k L_k^\top$. The latent variables are then sampled via the reparameterization trick. Due to the block-diagonal structure, this can be performed independently for each concept:
\begin{equation}
z_k = \mu_k + L_k \epsilon_k, \quad \epsilon_k \sim \mathcal{N}(0, I)
\end{equation}
Or in global vector form: $z = \mu + L \epsilon$.

\subsection{Flow-based Exogenous Prior}
Previous CDRL methods typically use a standard isotropic Gaussian prior as exogenous variables, i.e. $p(n) = \mathcal{N}(\mathbf{0}, I)$. Although mathematically convenient, this assumption is overly restrictive. Real-world causal factors often exhibit complex distributional characteristics that cannot be adequately captured by a unimodal Gaussian. To address this, we propose the Flow-based Exogenous Prior for each concept. We model exogenous variables $n_k$ for each causal concept $k$ using an independent Normalizing Flow, allowing the model to capture complex, non-Gaussian densities specific to each semantic factor.

We employ a set of independent Normalizing Flow~\cite{normalizingflow} models $\{T_{\psi,k}\}_{k=1}^K$ to define flexible densities for exogenous variables. Let $\epsilon_k \sim \mathcal{N}(\mathbf{0}, I)$ be a base distribution for the $k$-th concept. The flow defines a bijective mapping $n_k = T_{\psi,k}^{-1}(\epsilon_k)$. The log-likelihood of a noise sample $n_k$ is given by the change of variables formula:
\begin{equation}
\log p_{\psi,k}(n_k) = \log p(\epsilon_k) + \log \left| \det \frac{\partial T_{\psi,k}^{-1}(\epsilon_k)}{\partial \epsilon_k} \right|.
\end{equation}
Recall the structural equation for the $k$-th concept: $n_k = z_k - f_k(\{z_j \mid A_{jk}=1\})$. Due to the independence of exogenous noise terms, the joint likelihood is $\log p_{\text{SCM}}(z) = \sum_{k=1}^K \log p_{\text{SCM}}(z_k).$ Applying the change of variables from $n_k$ to $z_k$, one has:
\begin{equation}
\log p_{\text{SCM}}(z_k) = \log p_{\psi,k}(n_k) + \log \left| \det \frac{\partial n_k}{\partial z_k} \right|.
\end{equation}
Using Proposition~\ref{vpt}, the Jacobian term vanishes. This yields a computationally efficient objective for the prior:
\begin{equation}
\label{eq:pscm}
\log p_{\text{SCM}}(z) = \sum_{k=1}^K \log p_{\psi,k}\left( z_k - f_k({z_j \mid A_{jk}=1}) \right).
\end{equation}
This factorization allows us to decouple the learning of the causal mechanism from the learning of the distributional statistics. The model can then fit arbitrary complex exogenous distributions for each concept individually.

\subsection{Consistency Loss}
To ensure the decoder faithfully renders these causal interventions without introducing spurious correlations, we propose a Counterfactual Consistency Mechanism. This mechanism constrains the generative process by enforcing structural invariance.
\begin{definition}[Causal Partition]
    Given the causal graph $\mathcal{G}$, for a specific intervention on the latent variable $z_k$, the latent space $\mathbf{Z}$ is partitioned into three disjoint subsets\\
        \textbf{(1) Intervention Set ($\mathcal{I}_k$):} The singleton set containing the target variable itself, i.e., $\mathcal{I}_k = \{z_k\}$.\\
        \textbf{(2) Descendant Set ($\mathcal{D}_k$):} The set of variables strictly downstream from $z_k$, defined as $\mathcal{D}_k = \{z_j \in \mathbf{Z} \mid z_k \to \dots \to z_j \text{ in } \mathcal{G}\}$.\\
        \textbf{(3) Invariant Set ($\mathcal{S}_k$):} The complement set containing variables structurally independent of the intervention, defined as $\mathcal{S}_k = \mathbf{Z} \setminus (\mathcal{I}_k \cup \mathcal{D}_k)$.\\
\end{definition}
Based on the proposed partition, we stipulate that any intervention on $z_k$ must satisfy three key constraints. Specifically, interventional alignment for $\mathcal{I}_k$ across the decode and re-encode cycle, structural propagation for $\mathcal{D}_k$ governed by the SCM, and strict invariance for the invariant set $\mathcal{S}_k$.
Given a counterfactual latent code $\tilde{z}$ (derived by intervening on $z_k$ and fully propagating effects through the SCM), we decode it into an image $x_{\text{cf}}$ and re-encode it to obtain $\hat{z}$. The total consistency loss is formulated as:\begin{equation}
\begin{aligned}
\mathcal{L}_{\text{cons}}
&= \lambda_1 \| \hat{z}_{\mathcal{I}_k} - \tilde{z}_{\mathcal{I}_k} \|^2 + \lambda_2 \sum_{j \in \mathcal{D}_k} \| \hat{z}_j - \tilde{z}_j \|^2 \\
&+ \lambda_3 \sum_{j \in \mathcal{S}_k} \| \hat{z}_j - z_j \|^2,
\end{aligned}
\end{equation}
where $z_j$ is the $j$-th factor of the latent code $z$, $\tilde{z}_j$ is the ideal counterfactual value dictated by the causal graph, and $z_j$ is the original source value.

\subsection{Optimization Objective}
The Flow-based Prior is integrated into the Block-Diagonal VAE framework by modifying the KL divergence term in the Evidence Lower Bound (ELBO). We employ the Monte Carlo estimation for the KL divergence instead of an analytical computation, as the integration of the Flow-based prior's log-likelihood with the encoder's variational density yields no closed-form solution. Given a batch of posterior samples $z \sim q_\phi(z|x)$ from the Block-Diagonal Encoder, the prior matching loss is defined as:
\begin{equation}
\begin{aligned}
\mathcal{L}_{\text{KL}} &= \mathbb{E}_{q_\phi(z|x)} \left[ \log q_\phi(z|x) - \log p_{\text{SCM}}(z) \right] \\
&= \mathbb{E}_{q_\phi(z|x)} \left[\log q_\phi(z|x) - \sum_{k=1}^K \log p_{\psi,k}\left(n_k \right) \right].
\end{aligned}
\end{equation}
Intuitively, this objective encourages the encoder to produce latent codes $z$ such that, after removing the causal effects of parents, the remaining residuals $n$ follow a distribution that can be explained by the Flow model. This allows the framework to accommodate distributionally robust causal disentanglement respresentation learning, properly identifying factors even when they follow highly non-Gaussian statistics (e.g., Laplace or Bimodal), as validated in our experiments.
To learn the model parameters, we maximize the Evidence Lower Bound (ELBO) on the marginal likelihood. Substituting our Flow-based SCM prior into the KL divergence term, we obtain the following tractable training objective:
\begin{equation}
\text{ELBO}
= \mathbb{E}_{q_\phi(z|x)} [\log p_\theta(x|z)] - \beta \mathcal{L}_{KL}.
\end{equation}
Since encoder $q_\phi(z|x)$ infers latent variables solely from images, we introduce a supervision loss to explicitly align the learned causal blocks $z_k$ with the ground-truth $u_k$. This ensures that each subspace $z_k$ semantically encodes the specific factor designated by $u_k$:
\begin{equation}
\mathcal{L}_{\text{sup}} = \mathbb{E}_{q_\phi(z|x)} \left[ \sum_{k=1}^K \| h_k(z_k) - u_k \|^2 \right],
\end{equation}
where $h_k(\cdot)$ is a linear auxiliary predictor that maps the vector-valued $z_k$ to the label space of $u_k$.

To enforce the mutual independence derived from Assumption~\ref{ass:dependent}, we introduce the Hilbert-Schmidt Independence Criterion (HSIC)~\cite{hsic} loss $\mathcal{L}_{\text{HSIC}}$. This term explicitly minimizes the statistical dependence among exogenous variables during training. Finally, the total loss function $\mathcal{L}_{\text{total}}$ is as follows:
\begin{equation}
 \mathcal{L}_{\text{total}} = \mathcal{L}_{\text{recon}} + \beta \cdot \mathcal{L}_{\text{KL}} + \lambda \cdot \mathcal{L}_{\text{cons}} + \gamma \cdot \mathcal{L}_{\text{sup}} + \nu \cdot \mathcal{L}_{\text{HSIC}},   
\end{equation}
where $\beta, \gamma, \lambda,$ and $\nu$ are hyperparameters. The detailed derivation is provided in Appendix~\ref{loss}.

\subsection{Manifold-Aware Directional Intervention}
Although the hard intervention ($do(z_k=c)$) employed in previous methods is theoretically grounded, the inherent sparsity of the learned latent space often renders such forced scalar assignments prone to projecting latent codes off the data manifold, leading to out-of-distribution decoding and unrealistic generation. To mitigate this, we propose a Manifold-Aware Directional Intervention mechanism. This mechanism treats intervention as a dynamic traversal along a semantic vector field rather than a static assignment, effectively constraining the manipulation path within high-density data regions. Specifically, we capture a global direction $v_k \in \mathbb{R}^{d_k}$ representing the natural semantic variation of each factor $k$ (e.g., the transition from "Young" to "Old") via a momentum-based estimation strategy during training. By identifying sample subsets corresponding to the Top-K and Bottom-K ground-truth values within each batch, we continuously update their respective global expectations, $\mu_k^{\text{pos}}$ and $\mu_k^{\text{neg}}$, via Exponential Moving Average (EMA). The vector difference between these expectations is ultimately established as the robust direction for natural semantic variation.
\begin{equation}
\begin{aligned}
\mu_k^{\text{pos}} \leftarrow (1-\alpha)\mu_k^{\text{pos}} + \alpha \cdot \mathbb{E}_{z \in \mathcal{B}_{top}}[z_k],\\
 \quad \mu_k^{\text{neg} }\leftarrow (1-\alpha)\mu_k^{\text{neg}} + \alpha \cdot \mathbb{E}_{z \in \mathcal{B}_{btm}}[z_k],   
\end{aligned}
\end{equation}
where $\alpha$ is a momentum coefficient. $\mathcal{B}_{\text{top}}^{(k)}$ and $\mathcal{B}_{\text{btm}}^{(k)}$ denote the subsets of latent variables corresponding to the data points with the highest and lowest values of the ground-truth label $u_k$, respectively. The direction of intervention is defined as $\mathbf{v}_k = \mu_k^{\text{pos}} - \mu_k^{\text{neg}}$ and the Manifold-Aware Directional Intervention is then formalized as follows:
\begin{equation}
do(z_{k} = z_{k} + \tau \cdot \mathbf{v}_k),
\end{equation}
where $\tau$ is a scalar intensity. This fosters geometric alignment between the intervention path and the intrinsic data manifold.
\begin{table*}[t] 
\centering
\caption{Quantitative Comparison on Synthetic and Real-World Datasets. Note that the results of CausalDiffAE and CIDiffuser on Pendulum dataset are quoted from~\cite{cidiffuser}.}
\label{tab:main_results}
\resizebox{0.9\textwidth}{!}{
\begin{tabular}{l|ccc|cc|cc|cc}
\toprule
\multirow{2}{*}{\textbf{Method}} & \multicolumn{3}{c|}{\textbf{Filter}} & \multicolumn{2}{c|}{\textbf{Pendulum}} & \multicolumn{2}{c}{\textbf{CelebA (Smile)}} & \multicolumn{2}{c}{\textbf{CelebA (Age)}} \\
\cmidrule(lr){2-4} \cmidrule(lr){5-6} \cmidrule(lr){7-8} \cmidrule(lr){9-10} 
 & MIC ($\uparrow$) & TIC ($\uparrow$) & WD ($\downarrow$) & MIC ($\uparrow$) & TIC ($\uparrow$) & MIC ($\uparrow$) & TIC ($\uparrow$) & MIC ($\uparrow$) & TIC ($\uparrow$) \\
\midrule
\midrule
CausalDiffAE  & - & - & - & 91.10 \scriptsize{$\pm$0.00} & 89.20 \scriptsize{$\pm$0.00} & - & - & - & -\\
CIDiffuser    & - & - & - & 98.10 \scriptsize{$\pm$0.00} & 86.50 \scriptsize{$\pm$0.00} & - & - & - & -\\
SCMVAE     & 89.39 \scriptsize{$\pm$1.25}  & 63.20 \scriptsize{$\pm$0.33} & 89.29 \scriptsize{$\pm$0.40} & 81.47 \scriptsize{$\pm$0.42} & 76.40 \scriptsize{$\pm$1.48} & 52.61 \scriptsize{$\pm$0.62} & 43.65 \scriptsize{$\pm$0.58} 
& 38.10 \scriptsize{$\pm$0.61} & 29.71 \scriptsize{$\pm$0.57} \\
CausalVAE  & 95.61 \scriptsize{$\pm$0.20} & 76.87 \scriptsize{$\pm$0.05} & 82.40 \scriptsize{$\pm$0.09} & 91.06 \scriptsize{$\pm$1.21} & 82.08 \scriptsize{$\pm$0.39} & 57.65 \scriptsize{$\pm$0.36} & 48.51 \scriptsize{$\pm$0.21}
& \textbf{41.78} \scriptsize{$\pm$0.56} & 33.04 \scriptsize{$\pm$0.53}\\
\midrule
\textbf{FlexCausal (Ours)} & \textbf{98.88} \scriptsize{$\pm$0.10} & \textbf{89.54} \scriptsize{$\pm$0.02} & \textbf{57.99} \scriptsize{$\pm$0.40} & \textbf{98.94} \scriptsize{$\pm$0.49} & \textbf{90.13} \scriptsize{$\pm$0.53} & \textbf{60.26} \scriptsize{$\pm$0.58} & \textbf{54.86} \scriptsize{$\pm$0.55} & 41.56 \scriptsize{$\pm$0.32} & \textbf{37.23} \scriptsize{$\pm$0.22} \\
\bottomrule
\end{tabular}
}
\end{table*}

\section{Experiment}

\subsection{Experimental Setup}
\subsubsection{Datasets.}
\textbf{Filter} is a synthetic benchmark created using Blender~\cite{blender} to evaluate causal representation learning under challenging distributional constraints. The dataset consists of 12,000 images at $128 \times 128 \times 4$ resolution, generated by 6 ground-truth latent causal variables. The underlying causal structure is defined as: (filter size, filter position) $\to$ (shadow size), and (filter color, background color) $\to$ (shadow color). Distinctively, the generative factors follow bimodal and multimodal Gaussian distributions, creating a challenging disjoint latent structure for representation learning. Detailed information is provided in the Appendix~\ref{detail}.

\textbf{Pendulum}~\cite{causalvae} is a synthetic dataset which consists of four causal factors. The underlying causal structure is defined as: (Pendulum Angle, Light Position) $\to$ (Shadow Length, Shadow Position). 

\textbf{CelebA}~\cite{celeba} is a large-scale face attributes dataset containing more than 200k celebrity images with 40 binary attribute annotations. Specifically, we construct two distinct subsets, CelebA-Smile and CelebA-Age, to instantiate different causal mechanisms. CelebA-Smile focuses on facial expression dependencies. The underlying causal graph is defined as: (Smile, Gender) $\to$ (Narrow Eyes), (Smile) $\to$ (Mouth Open), and (Mouth Open) $\to$ (Narrow Eyes). CelebA-Age models demographic causal effects. The causal structure is defined as: (Age, Gender) $\to$ (Bald, Beard). For implementation, we randomly select 30,000 images from the dataset to constitute our training and testing sets.

\subsubsection{Metrics.}
We employ three complementary metrics to quantitatively assess the quality of learned causal representations. 
To evaluate identifiability and capture complex dependencies, we utilize the Mean Correlation Coefficient (MIC) and Total Information Coefficient (TIC)~\cite{mic}. Both measure the degree of information correlation between latent variables and ground truth.
To validate the effectiveness of our Flow-based prior, we compute the Wasserstein Distance (WD)~\cite{wd} between latent variables and ground-truth distributions.

\subsubsection{Implementation Details.}
We implemented FlexCausal using the PyTorch~\cite{paszke2019pytorch} framework on a single NVIDIA RTX 4090 GPU. For the backbone architecture, we employed a CNN as the encoder. To facilitate structural disentanglement, we utilized an independent additive decoder. The exogenous prior is parameterized by a stack of 4 Masked Autoregressive Flow (MAF) blocks~\cite{maf} to enable flexible density estimation. We took AdamW~\cite{adamw} as the optimizer with an initial learning rate of $1 \times 10^{-3}$. To ensure stable convergence, we adopted a cosine annealing scheduler paired with a linear warmup strategy. Details of network architectures and hyperparameter settings are provided in Appendix~\ref{detail}

\subsection{Experimental Results} 
We compare our model with SCM-VAE and CausalVAE across all datasets (Filter, CelebA-Smile, CelebA-Age, and Pendulum). Additionally, on the Pendulum dataset, we extend the comparison to include recent diffusion-based approaches, CIDiffuser and CausalDiffAE, to further evaluate our performance against generative diffusion baselines.

Quantitative results demonstrate that FlexCausal achieves the best performance on MIC, TIC, and WD metrics across all benchmarks. On Filter and CelebA datasets, FlexCausal outperforms SCM-VAE and CausalVAE, particularly in terms of MIC and TIC, validating its robust causal disentanglement and identifiability. Notably, on the Filter dataset, FlexCausal surpasses other methods in terms of WD. This confirms that our Flow-based Prior effectively captures complex ground-truth distributions comparable to other frameworks while maintaining the explicit structural control of SCMs.

\begin{table}[h]
\centering
\caption{Ablation Study on Filter Dataset.}
\label{tab:ablation}
\resizebox{\columnwidth}{!}{
\begin{tabular}{l|c|c|c}
\toprule
\textbf{Metrics} & MIC ($\uparrow$) & TIC ($\uparrow$) & WD ($\downarrow$) \\
\midrule
\textbf{FlexCausal (Ours)} & \textbf{98.88} \scriptsize{$\pm$0.10} & \textbf{89.54} \scriptsize{$\pm$0.02} & \textbf{57.99} \scriptsize{$\pm$0.40} \\
\midrule
w/o Flow Prior       & 91.88 \scriptsize{$\pm$2.07} & 75.81 \scriptsize{$\pm$2.36} & 60.29 \scriptsize{$\pm$1.96} \\
w/o Consistency Loss & 97.41 \scriptsize{$\pm$0.32} & 88.31 \scriptsize{$\pm$1.31} & 58.31 \scriptsize{$\pm$0.28}\\
w/o Block-Diagonal   & 98.19 \scriptsize{$\pm$0.19} & 89.05 \scriptsize{$\pm$0.03}  & 58.50 \scriptsize{$\pm$0.47} \\
\bottomrule
\end{tabular}
}
\end{table}

\subsection{Ablation Study}
As shown in Table~\ref{tab:ablation}, to evaluate the efficacy of each core component, we conducted an ablation study on the Filter dataset. Removing the Flow-based Prior yields the most significant performance degradation, marked by a drop in the TIC score and a significant deterioration in WD. This empirically confirms that standard Gaussian priors fail to capture complex exogenous noise, underscoring the indispensability of the flow model for precise structural identification and distribution matching. Furthermore, the exclusion of the Counterfactual Consistency Loss leads to declines in both MIC and TIC, validating the critical role of explicit structural penalties in ensuring robust causal disentanglement. Although removing the block diagonal mechanism yields only minor effects, the complete model demonstrates superior performance across all metrics, providing compelling evidence that the synergistic interaction of these components is indispensable for achieving robust causal generation.

\begin{figure}[!h]
    \centering
    \includegraphics[width=\columnwidth]{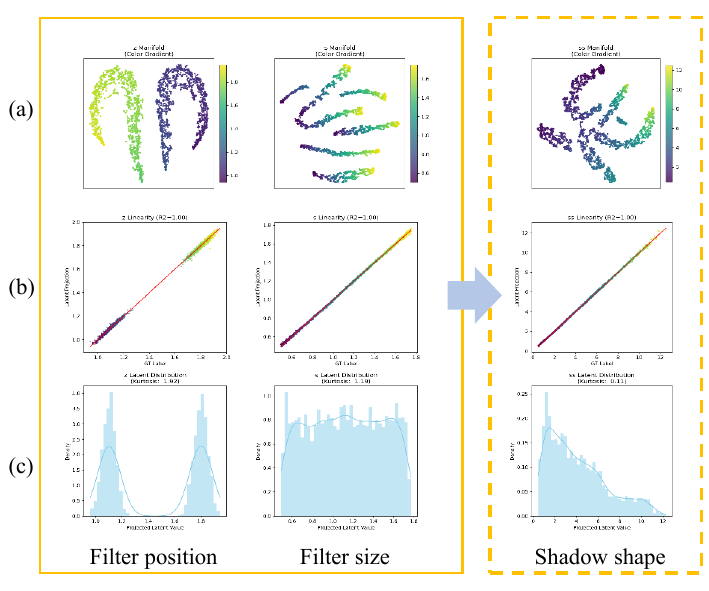}
    \caption{Visualization of the Learned Latent Space on the Filter Dataset.
(a) Latent Manifold Topology: The t-SNE visualization reveals the manifold structure of latent space.
(b) GT-Latent Alignment: The scatter plots demonstrate a strong linear correlation between latent projections and ground truth labels (with $R^2 > 0.99$ for almost every concept), confirming precise semantic alignment.
(c) Posterior Density: The learned latent distribution closely matches the ground truth distribution of the underlying factors, verifying the expressive capability of our Flow-based prior to model complex non-Gaussian densities.}
    \label{fig:distribution}
\end{figure}

\subsection{Visualization on Complex Distributions}
To verify FlexCausal's capability in modeling non-Gaussian and complex distributions, we visualize the learned latent space on the Filter dataset.
As illustrated in Figure~\ref{fig:distribution}, we project the high-dimensional latent variables into 2D/1D spaces to analyze their geometric and statistical properties from three distinct perspectives.
We first employ t-SNE~\cite{tsne} to visualize the global manifold structure of each latent block. As observed in Figure~\ref{fig:distribution}(a), FlexCausal effectively captures the intrinsic topology of the data, forming a smooth and continuous manifold.
Second, to assess disentanglement, we perform a linear regression between latent variables and GT. Figure~\ref{fig:distribution}(b) exhibits a strong linear correlation, with determination coefficients reaching near perfection ($R^2 > 0.99$) in almost all concepts. This indicates that FlexCausal achieves a near-isometric mapping, effectively disentangling semantic factors into orthogonal directions.
Third, the posterior density in Figure~\ref{fig:distribution}(c) demonstrates that the learned latent distribution highly aligns with the true underlying distribution of the GT factors (e.g. bimodal). This validates that our Flow-based prior successfully captures the authentic statistical properties of the data, avoiding both posterior collapse and the bias typically introduced by standard Gaussian regularization.

\section{Conclusions}
In this paper, we proposed FlexCausal, a novel CDRL framework designed to overcome the limitations of rigid Gaussian assumptions in Causal Disentangled Representation Learning. By synergizing a Block-Diagonal VAE with the flexible Flow-based Exogenous Prior, our approach effectively models complex non-Gaussian distributions while preserving the semantic integrity of vector-valued concepts. Furthermore, we introduced a Manifold-Aware Directional Intervention strategy to ensure structurally consistent and physically plausible counterfactual generation.
Extensive experiments on our proposed Filter benchmark and standard real-world datasets demonstrate that FlexCausal outperforms other CRDL methods.

\nocite{langley00}

\bibliography{flowprior}
\bibliographystyle{icml2026}

\newpage
\appendix
\onecolumn
\section{Derivation of Total Objective}
\label{loss}
\subsection{Derivation of the ELBO Objective}
Our goal is to maximize the marginal log-likelihood of the observed data $x$, denoted as $\log p_\theta(x)$. Since the true posterior $p_\theta(z|x)$ is intractable, we introduce a variational distribution $q_\phi(z|x)$ to approximate it. Starting from the log-likelihood: 
\begin{equation}
\begin{aligned}
\log p_\theta(x) &= \log \int p_\theta(x, z) \, dz \\
&= \log \int q_\phi(z|x) \frac{p_\theta(x, z)}{q_\phi(z|x)} \, dz \\
&= \log \mathbb{E}_{z \sim q_\phi(z|x)} \left[ \frac{p_\theta(x, z)}{q_\phi(z|x)} \right].
\end{aligned}    
\end{equation}

Applying Jensen's Inequality, we obtain the Evidence Lower Bound (ELBO):
\begin{equation}
\begin{aligned}
\log p_\theta(x) &\geq \mathbb{E}_{z \sim q_\phi(z|x)} \left[ \log \frac{p_\theta(x, z)}{q_\phi(z|x)} \right] \\
&= \mathbb{E}_{z \sim q_\phi(z|x)} \left[ \log p_\theta(x|z) + \log p_{\text{SCM}}(z) - \log q_\phi(z|x) \right] \\
&= \underbrace{\mathbb{E}_{q_\phi(z|x)} \left[\log p_\theta(x|z)\right]}_{\text{Reconstruction Term}} - \underbrace{D_{\text{KL}}(q_\phi(z|x) \| p_{\text{SCM}}(z))}_{\text{Regularization Term}},
\end{aligned}   
\end{equation}
 where the reconstruction term is computed using the Mean Squared Error(MSE). The core challenge lies in computing the KL divergence term, given that our prior $p_{\text{SCM}}(z)$ is a complex implicit distribution defined by a Normalizing Flow, and our posterior $q_\phi(z|x)$ has a structured block-diagonal form. The KL divergence term can be expanded as follows:
\begin{equation}
\begin{aligned}
D_{\text{KL}}(q_\phi(z|x) | p_{\text{SCM}}(z)) &= \mathbb{E}_{z \sim q_\phi} \left[\log q_\phi(z|x) - \log p_{\text{SCM}}(z) \right].
\end{aligned}
\end{equation}
Below, we derive the computation for terms $\log q_\phi(z|x) $ and $\log p_{\text{SCM}}$ respectively.

\subsection{Computation of Variational Log-Density $\log q_\phi(z|x)$}
We assume the variational posterior factorizes over $K$ causal concept blocks, $q_\phi(z|x) = \prod_{k=1}^K \mathcal{N}(z_k; \mu_k, \Sigma_k)$, where each covariance matrix is parameterized by a lower triangular Cholesky factor $L_k$ such that $\Sigma_k = L_k L_k^\top$.

We reparameterize $z_k$ for each block using a standard Gaussian noise vector $\epsilon_k \sim \mathcal{N}(0, I)$:
\begin{equation}
    z_k = \mu_k + L_k \epsilon_k.
\end{equation}
The log-density of the Gaussian sample $z_k$ is given by:
\begin{equation}
    \log q_\phi(z_k|x) = -\frac{d_k}{2}\log(2\pi) - \frac{1}{2}\log |\Sigma_k| - \frac{1}{2}(z_k - \mu_k)^\top \Sigma_k^{-1} (z_k - \mu_k).
\end{equation}
By substituting the reparameterization equation ($z_k - \mu_k = L_k \epsilon_k$) and the covariance factorization ($\Sigma_k = L_k L_k^\top$), the quadratic term simplifies as follows:
\begin{equation}
\begin{aligned}
    (z_k - \mu_k)^\top \Sigma_k^{-1} (z_k - \mu_k) &= (L_k \epsilon_k)^\top (L_k L_k^\top)^{-1} (L_k \epsilon_k) \\
    &= \epsilon_k^\top L_k^\top (L_k^\top)^{-1} L_k^{-1} L_k \epsilon_k \\
    &= \epsilon_k^\top \epsilon_k = \|\epsilon_k\|^2.
\end{aligned}
\end{equation}
Recall that the log-determinant of the covariance matrix simplifies via the Cholesky factor:
\begin{equation}
    \log |\Sigma_k| = \log |L_k L_k^\top| = 2 \sum_{i=1}^{d_k} \log (L_k)_{ii}.
\end{equation}
Substituting the simplified quadratic term ($\|\epsilon_k\|^2$) and the log-determinant term back into the definition of the multivariate Gaussian log-density, we obtain the closed-form expression for block $k$:
\begin{equation}
\begin{aligned}
    \log q_\phi(z_k|x) &= -\frac{d_k}{2}\log(2\pi) - \frac{1}{2}\log |\Sigma_k| - \frac{1}{2}(z_k - \mu_k)^\top \Sigma_k^{-1} (z_k - \mu_k) \\
    &= -\frac{d_k}{2}\log(2\pi) - \sum_{i=1}^{d_k} \log (L_k)_{ii} - \frac{1}{2} \|\epsilon_k\|^2.
\end{aligned}
\end{equation}
Finally, summing over all $K$ causal concept blocks, the total variational log-density is:
\begin{equation}
    \log q_\phi(z|x) = \sum_{k=1}^K \left( -\frac{d_k}{2}\log(2\pi) - \sum_{i=1}^{d_k} \log (L_k)_{ii} - \frac{1}{2} \|\epsilon_k\|^2 \right).
\end{equation}
This final form is computationally efficient as it relies solely on the predicted diagonal elements of $L_k$ and the sampled noise vectors $\epsilon_k$.

\subsection{Computation of $\log p_{\text{SCM}}(z) $}
\label{app:prior_ll}
Recall Equation~\ref{eq:pscm}, the SCM prior likelihood $\log p_{\text{SCM}}(z) $ is formulated as:
\begin{equation}
\log p_{\text{SCM}}(z) = \sum_{k=1}^K \log p_{\psi,k}\left( z_k - f_k({z_j \mid A_{jk}=1}) \right).
\end{equation}

\subsection{KL divergence term Estimation}
Combining the derived entropy term above and the SCM prior likelihood, the KL divergence is estimated via Monte Carlo sampling over a batch of size $B$:
\begin{equation}
    D_{\text{KL}}(q_\phi(z|x) | p_{\text{SCM}}(z))  \approx \frac{1}{B} \sum_{b=1}^B \left( \log q_\phi(z^{(b)}|x^{(b)}) - \log p_{\text{SCM}}(z^{(b)}) \right),
\end{equation}
where $z^{(b)}$ represents the reparameterized samples. This estimaton enables efficient end-to-end optimization of both the encoder and the SCM parameters.

\subsection{Final Training Objective}
\label{app:final_loss}
Integrating ELBO with the proposed causal constraints, the final differentiable loss function is defined as:
\begin{equation}
\mathcal{L}_{\text{total}} = 
\underbrace{ ||x - D_\theta(z)||^2 }_{\text{Reconstruction}} 
+ \beta \underbrace{ \left( \log q_\phi(z|x) - \log p_{\text{SCM}}(z) \right) }_{\text{KL Divergence}} 
+ \lambda \mathcal{L}_{\text{cons}} 
+ \gamma \mathcal{L}_{\text{sup}}
+ \nu \mathcal{L}_{\text{HSIC}},
\end{equation}
where constants are omitted for brevity, and $z$ is the reparameterized sample $z = \mu + L\epsilon$. Hyperparameters $\beta, \lambda, \gamma, \nu$ control the trade-off between reconstruction quality, distribution matching, and structural constraints.

\section{Proof of Proposition}
\subsection{Proof of Proposition~\ref{vpt}}
\begin{proof}
    Let $z = [z_1^\top, \dots, z_K^\top]^\top$ be the latent vector partitioned into $K$ causal blocks, where $z_k \in \mathbb{R}^{d_k}$ and the total dimension is $D = \sum_{k=1}^K d_k$. Let $\mathbf{J} \in \mathbb{R}^{D \times D}$ denote the Jacobian matrix of the transformation mapping the latent variables $z$ to the exogenous noise $n$, defined block-wise as $\mathbf{J}_{ij} = \frac{\partial n_i}{\partial z_j}$. Recall Equation~\ref{eq:anm}:
    \begin{equation}
        n_k = z_k - f_k(\text{PA}(z_k)).
    \end{equation}

    The diagonal block $\mathbf{J}_{kk}$ represents the partial derivative of $n_k$ with respect to $z_k$. Differentiating the structural equation yields:
    \begin{equation}
        \mathbf{J}_{kk} = \frac{\partial n_k}{\partial z_k} = \frac{\partial \left( z_k - f_k(\text{PA}(z_k)) \right)}{\partial z_k} = \mathbf{I}_{d_k} - \frac{\partial f_k(\text{PA}(z_k))}{\partial z_k},
    \end{equation}
    where $\mathbf{I}_{d_k}$ is the $d_k \times d_k$ identity matrix. Due to the acyclicity of the causal graph $\mathcal{G}$, a variable $z_k$ cannot be its own parent (i.e., $z_k \notin \text{PA}(z_k)$). $f_k(\text{PA}(z_k))$ is independent of $z_k$, implying $\frac{\partial f_k}{\partial z_k} = \mathbf{0}$. Thus, the diagonal blocks simplify to:
    \begin{equation}
        \mathbf{J}_{kk} = \mathbf{I}_{d_k}.
    \end{equation}
    Assume the indices $1, \dots, K$ follow a valid topological ordering of the graph $\mathcal{G}$. For any $j > i$, the variable $z_j$ implies a node downstream or unrelated to $z_i$, and thus cannot be a parent of $z_i$. Therefore, $n_i$ is independent of $z_j$, which implies:
    \begin{equation}
        \mathbf{J}_{ij} = \frac{\partial n_i}{\partial z_j} = \mathbf{0} \quad \text{for all } j > i.
    \end{equation}
    Finally, we obtain:
    \begin{equation}
        \det(\mathbf{J}) = \prod_{k=1}^K \det(\mathbf{J}_{kk}) = \prod_{k=1}^K \det(\mathbf{I}_{d_k}) = \prod_{k=1}^K 1 = 1.
    \end{equation}
    Since the Jacobian determinant is unity, we conclude that the transformation is volume-preserving.
\end{proof}

\section{Detailed Information}
\label{detail}
\subsection{Filter Dataset}
\begin{figure*}[!htbp]
    \centering
    \includegraphics[width=\textwidth]{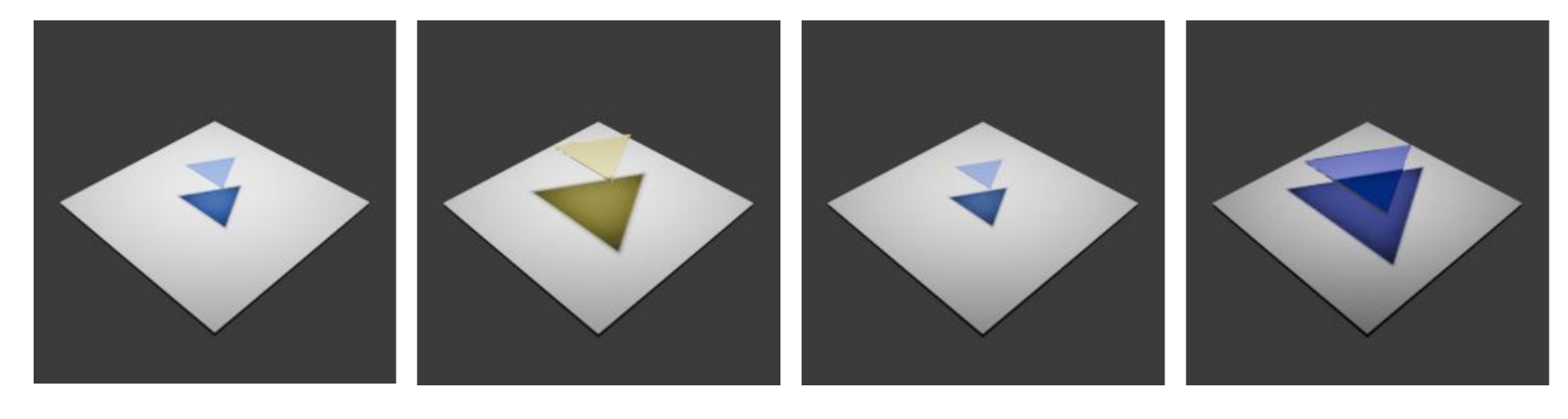}
    \caption{Filter Dataset}
    \label{fig:filter}
\end{figure*}

To evaluate the model's capability in handling complex physical interactions and non-standard latent distributions, we introduce the \textbf{Filter} dataset. As shown in Figure~\ref{fig:filter}, this dataset is synthetically generated using the Blender 3D creation suite with the Cycles physics-based rendering engine. The dataset contains 12,000 high-quality rendered images with a resolution of $128 \times 128 \times 4$ (RGBA). The inclusion of the alpha channel and the high-fidelity ray-traced shadows provides rich supervision for learning fine-grained causal mechanisms.

\textbf{Data Generation Process.} The scene consists of a translucent equilateral triangular filter, a fixed point light source , and a receiving surface. The dataset focuses on capturing the causal interplay between the geometry of the occluder and the resulting projection. The underlying Structural Causal Model (SCM) involves 4 independent exogenous variables (parents) and 2 dependent endogenous variables (children):
\begin{itemize}
    \item \textbf{Filter Size ($S$):} The side length of the equilateral triangle, sampled from a continuous uniform distribution.
    \item \textbf{Filter Position ($H$): }The vertical distance of the filter from the base, sampled from a bimodal distribution.
    \item \textbf{Filter Color ($C_f$): }The RGB absorption properties of the filter.
    \item \textbf{Background Color ($C_b$): }The albedo color of the ground surface.
    
    \item \textbf{Shadow Shape ($A_s$):}Determined by the interaction between $S$ and $H$ under the perspective projection of the point light ($A_s := f(S, H)$).
    \item \textbf{Shadow Color ($C_s$): }Resulting from the subtractive color mixing of the light passing through the filter onto the colored base ($C_s := g(C_f, C_b)$).
\end{itemize}

\textbf{Distributional Challenges.} Unlike standard datasets that often assume unimodal Gaussian latents, CausalFilter is specifically designed to challenge representation learning algorithms by enforcing bimodal and multimodal Gaussian distributions on key causal variables. Specifically, the Filter Position is sampled from two distinct modes ($z \in \{1.1, 1.8\}$ with Gaussian noise), and Filter Color is sampled from three distinct clusters (Red , Blue, and Yellow). This creates a discontinuous and disjoint manifold structure in the ground-truth latent space, rigorously testing the model's ability to learn disentangled representations and perform valid counterfactual generation across diverse modes.

\subsection{Implement Detail}
Table~\ref{tab:network_arch} details the specific architecture of the Encoder and Decoder used in our experiments. For the Flow-based prior, we utilized a Masked Autoregressive Flow (MAF) with 4 layers.

\begin{table}[h]
    \centering
    \caption{\textbf{Network Architecture of FlexCausal (CelebA 128x128).} The model consists of a shared encoder backbone and an independent additive decoder. Causal concepts and other concepts share the backbone but use distinct prediction heads. $K$ denotes the number of causal concepts. For simplicity, we employ a standard VAE to encode other concepts in CelebA that are not part of the SCM.}
    \label{tab:network_arch}
    \begin{tabular}{l l l}
    \toprule
    \textbf{Stage} & \textbf{Layer} & \textbf{Output Shape} \\
    \midrule
    \multicolumn{3}{c}{\textit{\textbf{Encoder}}} \\
    \midrule
    \multirow{5}{*}{Shared Backbone} 
        & Input Image & $3 \times 128 \times 128$ \\
        & Conv2d(3, 32, k=3, s=2), BN, LeakyReLU & $32 \times 64 \times 64$ \\
        & ResBlock(32, 64, s=2) & $64 \times 32 \times 32$ \\
        & ResBlock(64, 128, s=2) & $128 \times 16 \times 16$ \\
        & ResBlock(128, 256, s=2) & $256 \times 8 \times 8$ \\
    \midrule
    \multirow{2}{*}{Prediction Heads} 
        & Conv2d(256, 512, k=3, s=2), BN, LeakyReLU & $512 \times 4 \times 4$ \\
        & Flatten $\to$ Linear $\to$ Split($\mu, \sigma$) & $K \times z_{dim}$, $K \times z_{dim} \times z_{dim}$ \\
    \midrule
    \multicolumn{3}{c}{\textit{\textbf{Decoder}}} \\
    \midrule
    \multirow{2}{*}{Independent Projectors} 
        & Linear($z_{dim}$, 8192) $\to$ Reshape & $K \times (512 \times 4 \times 4)$ \\
        & \textbf{Additive Fusion:} $\sum_{k=1}^K \text{Feat}_k$ & $512 \times 4 \times 4$ \\
    \midrule
    \multirow{5}{*}{Upsampling Body} 
        & Upsample(2x), Conv(512, 256), ResBlock & $256 \times 8 \times 8$ \\
        & Upsample(2x), Conv(256, 128), ResBlock & $128 \times 16 \times 16$ \\
        & Upsample(2x), Conv(128, 64), ResBlock & $64 \times 32 \times 32$ \\
        & Upsample(2x), Conv(64, 32), ResBlock & $32 \times 64 \times 64$ \\
        & Upsample(2x), Conv(32, 3), Tanh & $3 \times 128 \times 128$ \\
    \bottomrule
    \end{tabular}
\end{table}

We list the hyperparameters used for training in Table~\ref{tab:hyperparams}.
\begin{table}[!htbp]
    \centering
    \caption{\textbf{Hyperparameter Settings.} The Hyperparameter configuration(CelebA).}
    \label{tab:hyperparams}
    \begin{tabular}{l c}
    \toprule
    \textbf{Hyperparameter} & \textbf{Value} \\
    \midrule
    Batch Size   & 64\\
    Total Epochs & 100 \\
    \midrule
    Normalizing Flow Layers & 4 \\
    Latent Dimension     & 16  \\
    \midrule
    Consistency Loss Weight ($\lambda_{1}$) & 1.0 \\
    Consistency Loss Weight ($\lambda_{2}$) & 10.0 \\
    Consistency Loss Weight ($\lambda_{3}$) & 1.0 \\
    \midrule
    KL Loss Weight ($\beta$) & 0.1 \\
    Sup Loss Weight ($\gamma$) & 3000 \\
    \bottomrule
    \end{tabular}
\end{table}

\newpage
\section{Additional Experimental Results}
\begin{figure*}[htbp]
    \centering
    \includegraphics[width=0.72\textwidth]{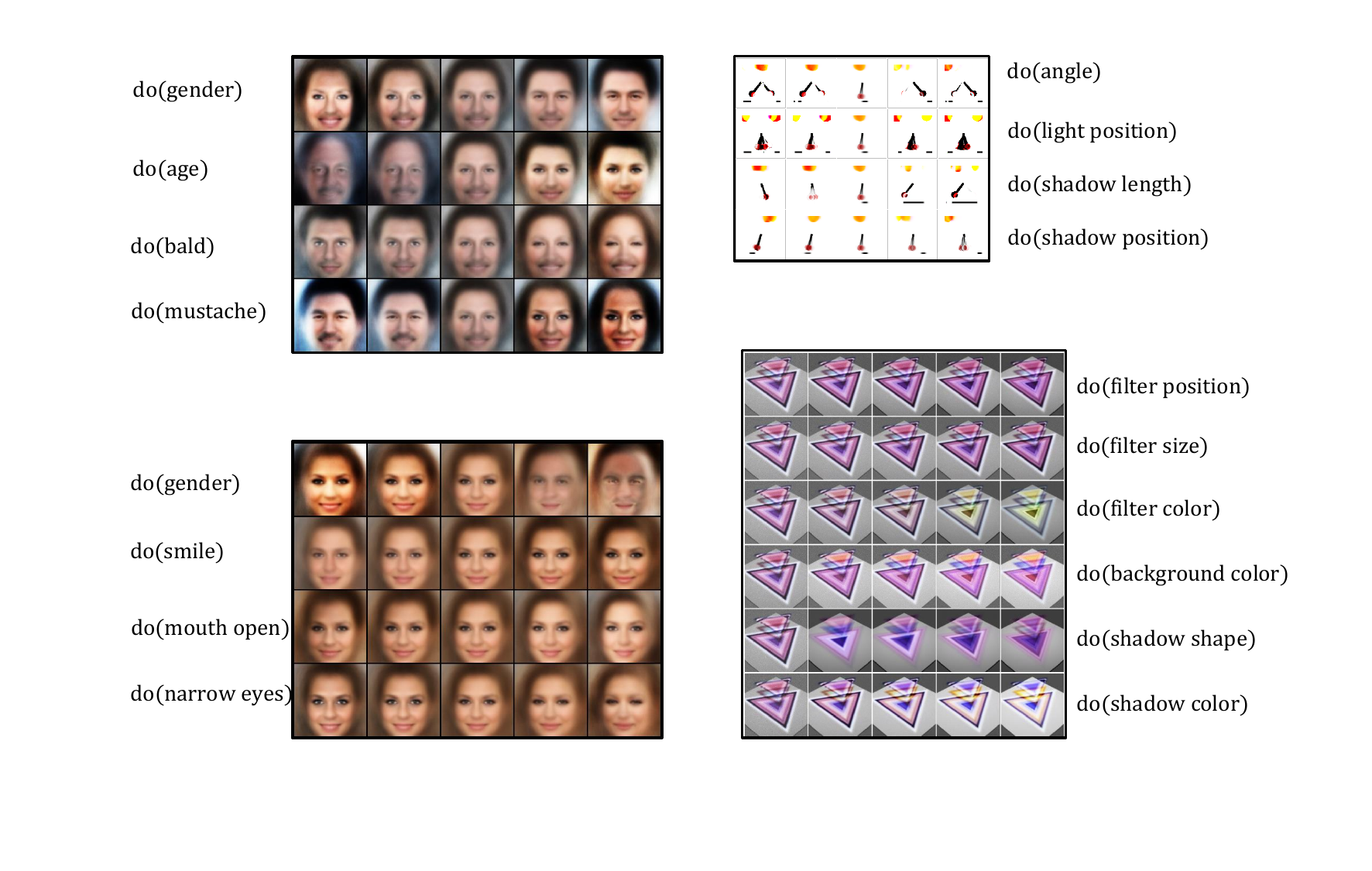}
    \caption{Resuls of CausalVAE on CelebA-Smile, CelebA-Age, Pendulum, Filter}
    \label{fig:placeholder}
\end{figure*}

\begin{figure*}[htbp]
    \centering
    \includegraphics[width=0.72\textwidth]{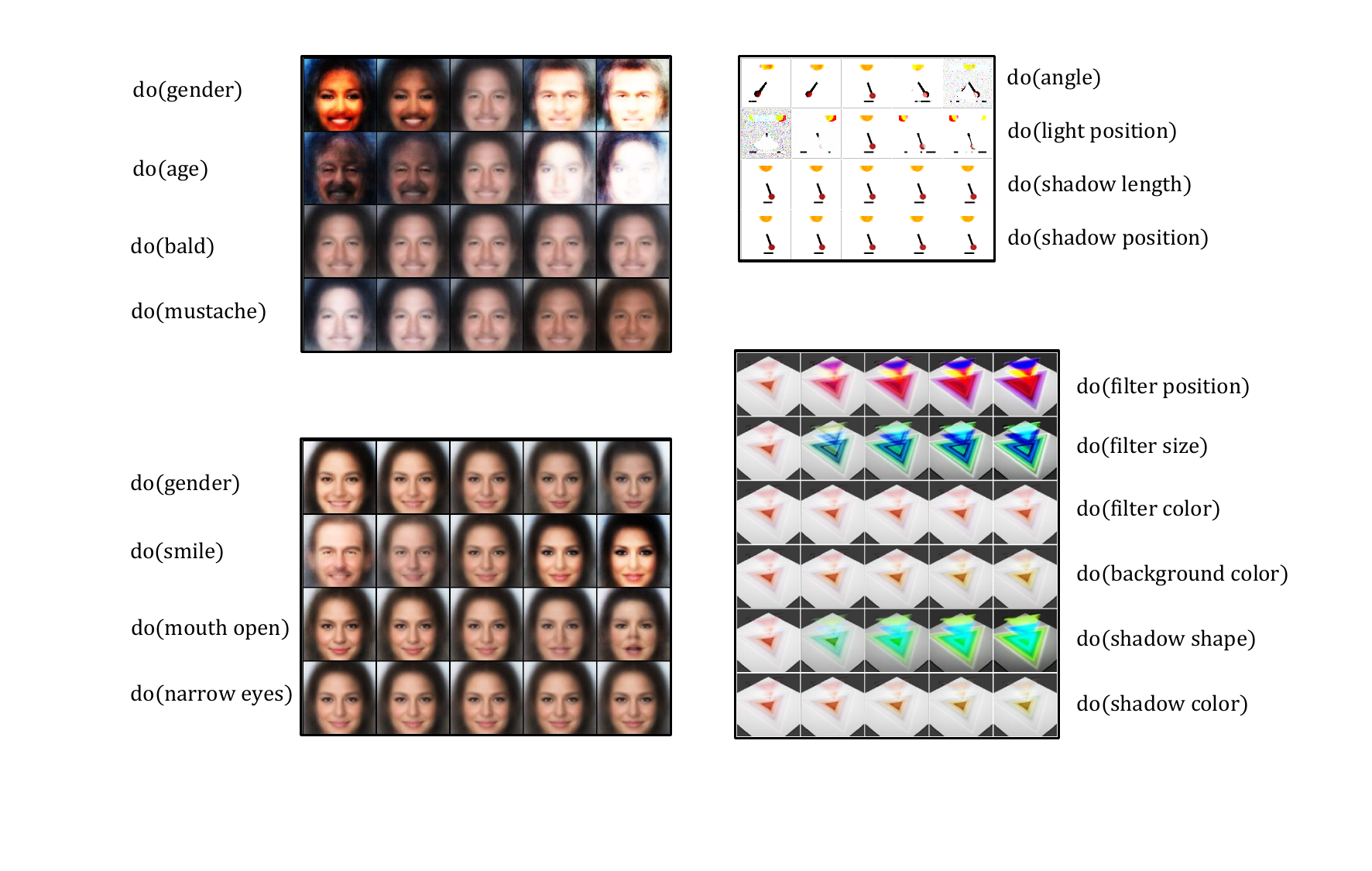}
    \caption{Resuls of SCMVAE on CelebA-Smile, CelebA-Age, Pendulum, Filter}
    \label{fig:placeholder}
\end{figure*}

\begin{figure*}[htbp]
    \centering
    \includegraphics[width=\textwidth]{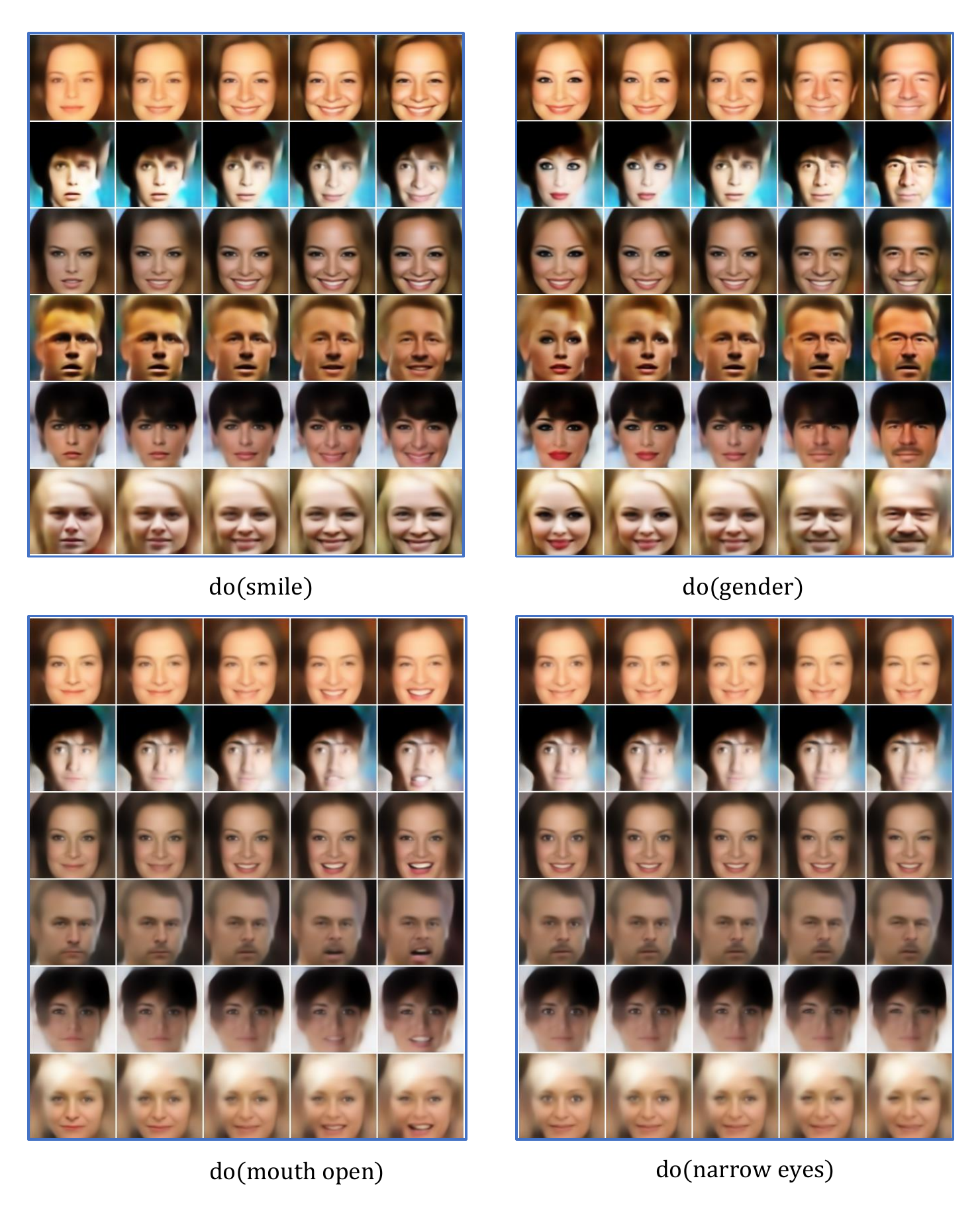}
    \caption{Resuls of our FlexCausal on CelebA-Smile}
    \label{fig:placeholder}
\end{figure*}

\begin{figure*}[htbp]
    \centering
    \includegraphics[width=\textwidth]{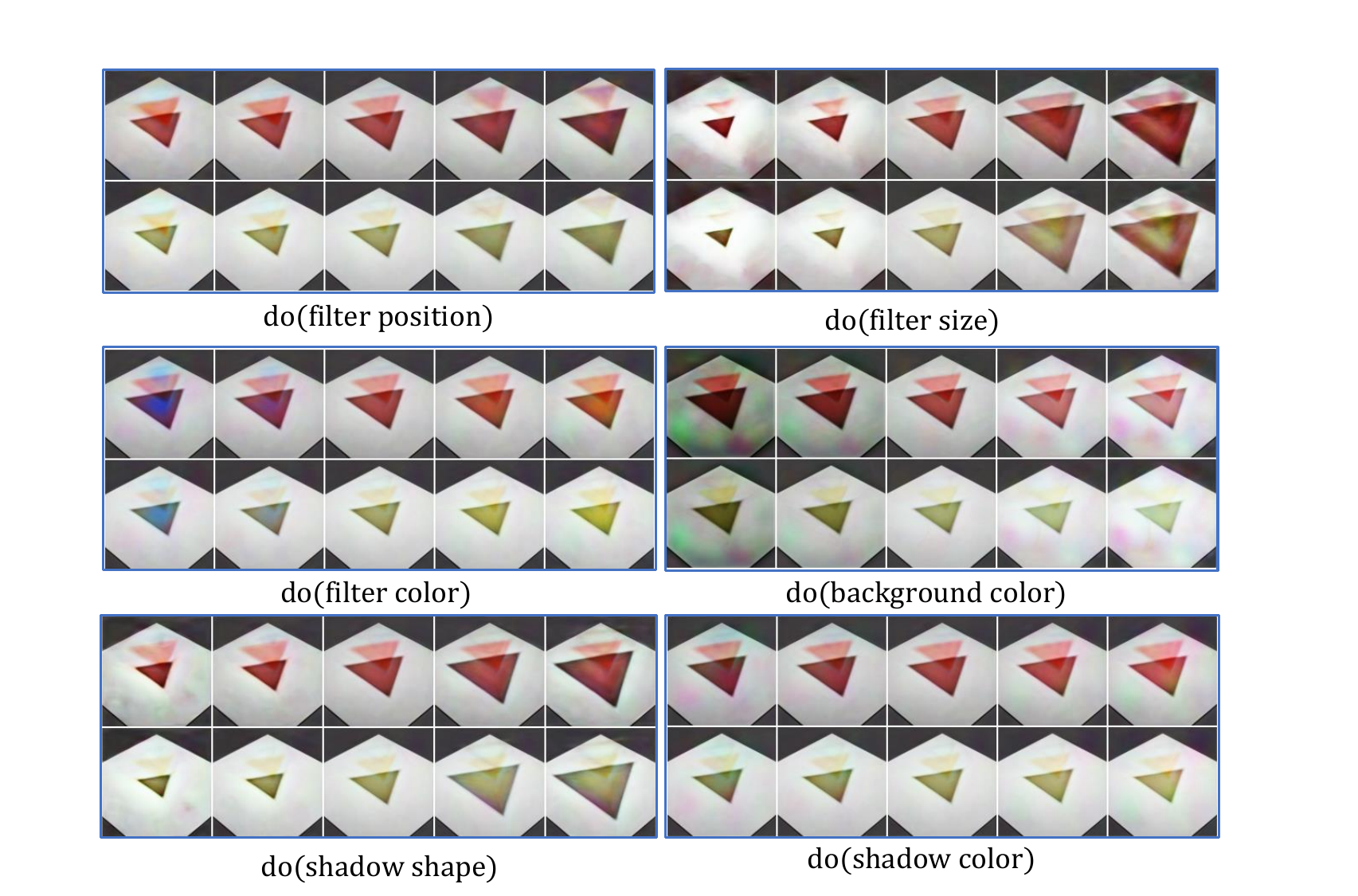}
    \caption{Resuls of our FlexCausal on Filter}
    \label{fig:placeholder}
\end{figure*}

\begin{figure*}[htbp]
    \centering
    \includegraphics[width=\textwidth]{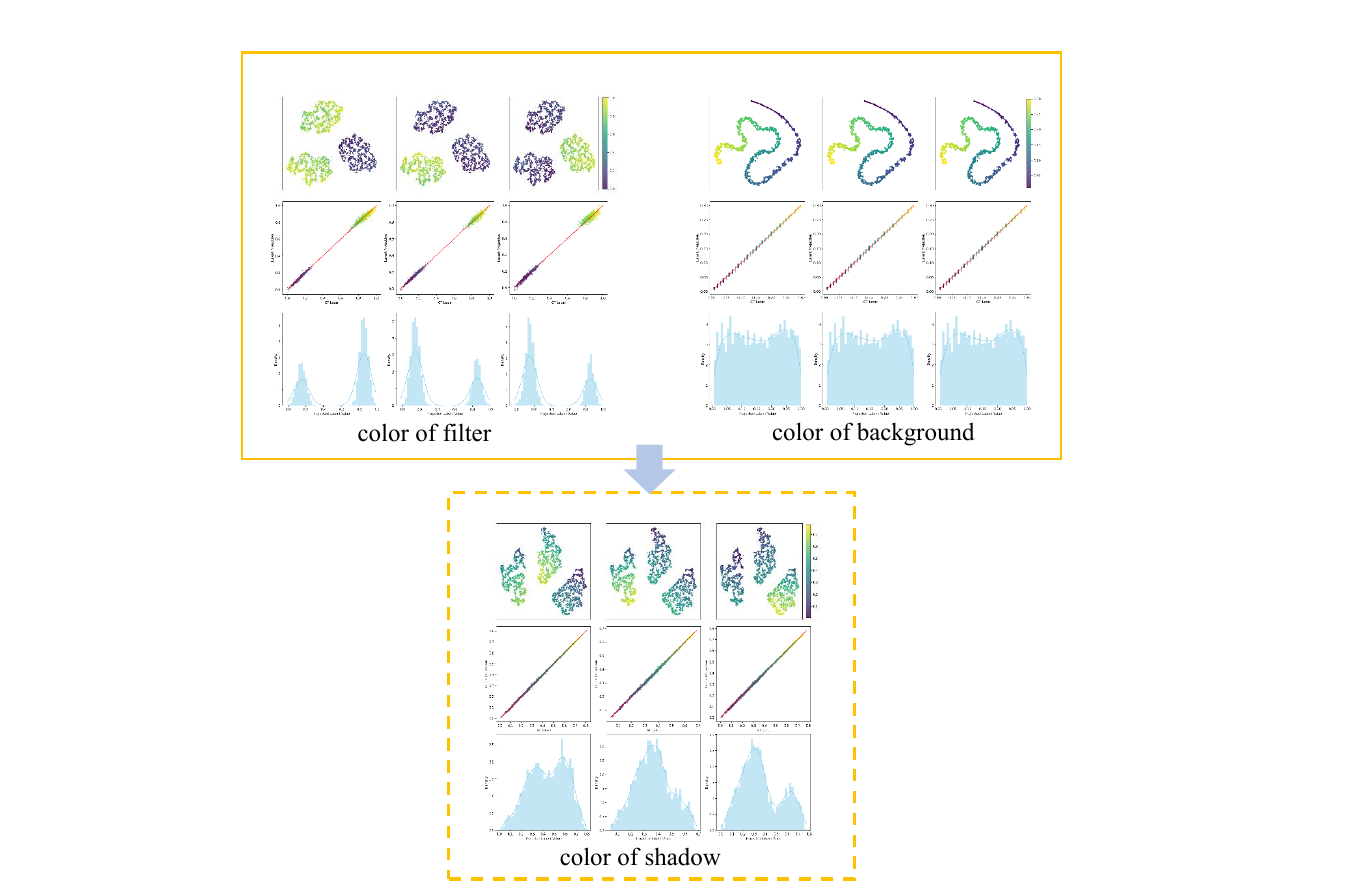}
    \caption{Visualization of Filter color, Background color, Shadow color}
    \label{fig:placeholder}
\end{figure*}

\newpage
\section{Algorithm}
\subsection{FlexCausal Pretrain Algorithm}
\begin{algorithm}[htbp]
\caption{Training Procedure for FlexCausal}
\label{alg:training}
\begin{algorithmic}[1]
\REQUIRE Dataset $\mathcal{D}=\{x^{(i)}, u^{(i)}\}_{i=1}^N$, Adjacency Matrix $A$, Momentum $\alpha$, Hyperparams $\beta, \lambda, \gamma, \nu$.
\STATE \textbf{Initialize:} Encoder $\phi$, Decoder $\theta$, SCM parameters $\omega$, Flow parameters $\psi$.
\STATE \textbf{Initialize:} Global manifold centroids $\mu_k^{\text{pos}}, \mu_k^{\text{neg}} \leftarrow \mathbf{0}$ for $k=1, \dots, K$.
\FOR{epoch$=1$ to $M$}
    \STATE Sample mini-batch $\{x, u\} \sim \mathcal{D}$.
    
    \STATE \textcolor{blue}{\textbf{// Block-Diagonal Encoding}}
    \STATE $\mu_\phi, \Sigma_\phi \leftarrow \text{Encoder}_\phi(x)$ 
    \STATE Sample latent variables: $z_k \leftarrow \mu_k + L_k \cdot \epsilon_k, \quad \epsilon_k \sim \mathcal{N}(0, I)$.

    \STATE \textcolor{blue}{\textbf{// Manifold-Aware Direction Update}}
    \FOR{$k=1$ to $K$}
        \STATE Identify Top/Bottom-K indices:
        \STATE $\mathcal{I}_{\text{top}} \leftarrow \text{Top-K Indices}(u_k)$; \quad $\mathcal{I}_{\text{btm}} \leftarrow \text{Bottom-K Indices}(u_k)$.
        \STATE Compute batch expectations:
        \STATE $\mathbb{E}_{\text{top}} \leftarrow \text{Mean}(z_k[\mathcal{I}_{\text{top}}])$; \quad $\mathbb{E}_{\text{btm}} \leftarrow \text{Mean}(z_k[\mathcal{I}_{\text{btm}}])$.
        \STATE Update global directions:
        \STATE $\mu_k^{\text{pos}} \leftarrow \alpha \mu_k^{\text{pos}} + (1-\alpha) \mathbb{E}_{\text{top}}$; $\mu_k^{\text{neg}} \leftarrow \alpha \mu_k^{\text{neg}} + (1-\alpha) \mathbb{E}_{\text{btm}}$
        \STATE Update inference vector: $\mathbf{v}_k \leftarrow \mu_k^{\text{pos}} - \mu_k^{\text{neg}}$
    \ENDFOR

    \STATE \textcolor{blue}{\textbf{// Flow-based Prior Evaluation}}
    \FOR{$k=1$ to $K$}
        \STATE Get parents: $\text{PA}(z_k) \leftarrow \{z_j \mid A_{jk}=1\}$.
        \STATE Compute residual: $n_k \leftarrow z_k - f_{\omega, k}(\text{PA}(z_k))$.
        \STATE Evaluate Flow likelihood: $\ell_k \leftarrow \log p(T_\psi(n_k)) + \log |\det J_{T_\psi}|$.
        \STATE $\log p_{\text{SCM}}(z) \leftarrow \log p_{\text{SCM}}(z) + \ell_k$
    \ENDFOR
    \STATE \textcolor{blue}{\textbf{// Counterfactual Consistency}}
    \STATE Construct $z_{\text{cf}}$: Set $z_{\text{cf}}^{(k)} \leftarrow \tilde{z}^{(k)}$, keep invariants $z_{\text{cf}}^{(\neg desc)} \leftarrow z^{(\neg desc)}$, and update children $z_{\text{cf}}^{(desc)}$.
    \STATE Re-encode: $\hat{z}_{\text{cf}} \leftarrow E_\phi(D_\theta(z_{\text{cf}}))$.
    
    \STATE \textcolor{blue}{\textbf{// Loss Computation}}
    \STATE $\mathcal{L}_{\text{KL}} \leftarrow \mathbb{E}_{q_\phi} [ \log q_\phi(z|x) - \log p_{\text{SCM}}(z) ]$
    \STATE $\mathcal{L}_{\text{recon}} \leftarrow -\log p_\theta(x|z)$.
    \STATE $\mathcal{L}_{\text{cons}} \leftarrow \|\hat{z}_{\text{cf}}^{(k)} - \tilde{z}^{(k)}\|^2+ \|\hat{z}_{\text{cf}}^{(\neg desc)} - z^{(\neg desc)}\|^2 + \|\hat{z}_{\text{cf}}^{(desc)} - z_{\text{cf}}^{(desc)}\|^2$.
    \STATE $\mathcal{L}_{\text{sup}} \leftarrow \sum_{k=1}^K \| h_k(z_k) - u_k \|^2$. 

    \STATE \textcolor{blue}{\textbf{// Optimization}}
    \STATE $\mathcal{L}_{\text{total}} \leftarrow \mathcal{L}_{\text{recon}} + \beta \mathcal{L}_{\text{KL}}  + \lambda \mathcal{L}_{\text{sup}} + \gamma \mathcal{L}_{\text{cons}} + \nu \mathcal{L}_{\text{HSIC}}$.
    \STATE Update parameters $\phi, \theta, \omega, \psi \leftarrow \text{Optimizer}(\nabla \mathcal{L}_{\text{total}})$.

\ENDFOR
\end{algorithmic}
\end{algorithm}


\end{document}